# A Narrative Vehicle Protection Representation for Vehicle Speed Regulator Under Driver Exhaustion-A Study

[1]V.Karthikeyan; [2]B.Praveen Kumar; [3]S.Suresh Babu;

[4]R.Purusothaman; [5]shijin Thomas

[1] Assistant Professor, Department of ECE, SVS College of Engineering, Coimbatore

[2,3,4,5] UG Students Department of ECE, SVS College of Engineering, Coimbatore

[1]Karthick77keyan@gmail.com

**ABSTRACT:**

*Driver fatigue is one of the important factors that cause traffic accidents, and the ever-increasing number due to diminished driver's vigilance level has become a problem of serious concern to society. Drivers with a diminished vigilance level suffer from a marked decline in their abilities of perception, recognition, and vehicle control, and therefore pose serious danger to their own life and the lives of other people. Exhaustion resulting from sleep deprivation or sleep disorders is an important factor in the creasing number of accidents. In this projected work, we discuss the various methods of the existing and the proposed method based on a real-time online safety prototype that controls the vehicle speed under driver fatigue. The purpose of such a model is to advance a system to detect fatigue symptoms in drivers and control the speed of vehicle to avoid accidents. This system was tested adequately with subjects of different technology of various researchers; finally the validity of the proposed model for vehicle speed controller based on driver fatigue detection is shown.*

## 1. INTRODUCTION:

Driver fatigue is one of the important factors that cause traffic accidents, and the ever-increasing number due to admonished driver's vigilance level has become a problem of serious concern to society. Drivers with a diminished vigilance level suffer from a marked decline in their abilities of perception, recognition, and vehicle control [6] causing death is fatigue-related [3]. Statistics show that a leading cause of fatal or injury-causing traffic accidents is due to driver fatigue. In the trucking industry, 57% of fatal truck accidents are due to with a diminished vigilance level. It is the number one cause of heavy truck crashes. Seventy percent of American drivers report driving fatigued. In commercial motor vehicle drivers, 44% took at least one nap during a duty cycle that contained clinically-scorable sleep [4]. The National Highway Traffic Safety Administration (NHTSA) estimates that there are 100 000 crashes that are caused by drowsy drivers and result in more than 1500 fatalities and 71 000 injuries each year in U.S. In China, the number of accident that caused by fatigue is the highest not only in absolute figure but also in proportion. With the ever-growing traffic conditions, this problem will further increase. Therefore, how to avoid fatigue driving efficiently can help prevent many accidents, consequently save money and reduce personal suffering. For this reason, developing driver fatigue detection systems [8] and related intelligent safety applications are very important to vehicular systems in the future. In the past ten years, many countries all over the world have begun to pay attention to the driver safety problem and to investigate the driver's mental states relating to driving safety. And some driver fatigue methods have been proposed which can detect whether the driver is tired, such as drowsy or inattention, for generating some warning alarms to alert the driver. All of these efforts have been reported in the literature for developing an active safety system for reducing the number of automobile accidents due to driver fatigue. Despite the success of the existing approaches/systems for extracting characteristics of a driver using computer vision technologies, current efforts in this area, it is a challenging issue due to a variety of factors. The first reason is the variety of eyes moving fast, external illuminations interference and realistic lighting conditions. So non-intrusive eye tracking [5] is a very difficult work in driving environments At the same time, it's very difficult to model the driver's eye movement dynamics because of the eye motion being the high nonlinearity.

## 2. RELATED WORKS:

**Anti-Locking Braking System** works with your regular braking system by automatically pumping them. In vehicles not equipped with ABS, the driver has to manually pump the brakes to prevent wheel lockup. In vehicles equipped with ABS, your foot should remain firmly planted on the brake pedal, while ABS pumps the brakes for you so you can concentrate on steering to safety. **Electronic brake-force distribution** Electronic brake-force distribution (EBD or EBFD) Electronic brake-force limitation (EBL) is an automobile brake technology that automatically varies the amount of force applied to each of a vehicle's brakes, based on road conditions, speed, loading, etc. always coupled with anti-lock braking systems. **Supplemental Restraint System Air Bags** An airbag is a vehicle safety device. It is an occupant restraint consisting of a flexible envelope designed to inflate rapidly during an automobile collision, to prevent occupants from striking interior objects such as the steering wheel or a window, the sensors may deploy one or more airbags in an impact zone at variable rates based on the type and severity of impact; the airbag is designed to only inflate in moderate to severe frontal crashes.

## 3. PROPOSED MODEL:

The various steps involved in the proposed model are discussed in the following components **Image acquisition and eye detection;** Image Acquisition System Image understanding of visual behaviors starts with image acquisition. The purpose of image acquisition is to acquire the video images of the driver face in real time. The acquired images should have relatively consistent photometric property under different climatic/ambient conditions and should produce distinguishable features that can facilitate the subsequent image processing. To this end, the person's face is illuminated using a near-infrared illuminator.**Eyelid Movement Parameters;** Eyelid movement [3] is one of the visual behaviours that reflect a person's level of fatigue. The primary purpose of eye tracking [3], [5]is to monitor eyelid movements and compute the relevant eyelid movement parameters. Here, we focus on two ocular measures to characterize the eyelid movement [3]

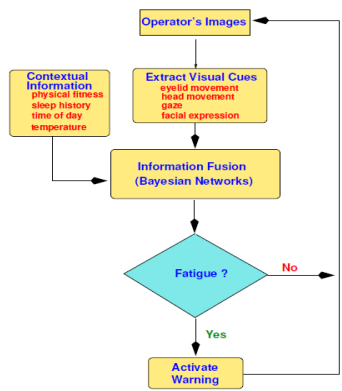

**Fig 1: flow chart for driver vigilance monitoring**

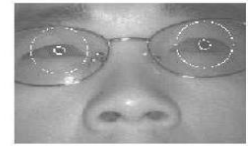

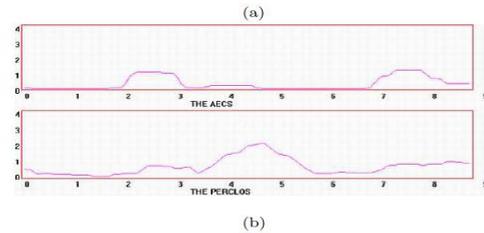

**Fig 2: a) Detected eyes and pupils b) plot for eyelid movement parameters**

The first one is Percentage of Eye Closure over Time (**PERC-LOS**) and the second is Average Eye Closure Speed (**AECS**). PERCLOS has been validated and found to be the most valid ocular parameter for monitoring fatigue. Eye tracking [5] starts with eyes detection. Figure gives a flowchart of the eye detection procedure. Eyes detection is accomplished via pupil's detection due to the use of active IR illumination.

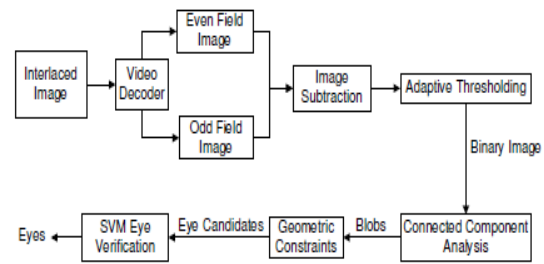

**Fig 3: Block diagram for driver vigilance monitoring**

The inner ring of LEDs and outer ring of LEDs with the even and odd fields of the interlaced image respectively so that they can be turned on and off alternately. The interlaced input image is de-interlaced via a video decoder, producing the even and odd field images as shown in Figure.

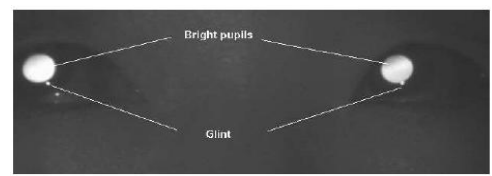

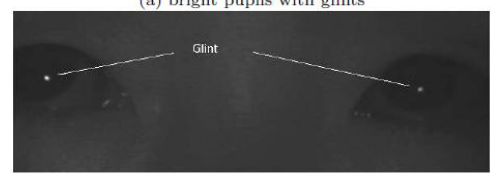

**Fig 4: Bright and Dark pupil images with glints**

The interlaced input image is de-interlaced via a video decoder, producing even and odd images as shown in Figure. While both images share the same background and external illumination, pupils in the even images look significantly brighter than in the odd images. To eliminate the background and reduce external light illumination, the odd image is subtracted from the even image, producing the difference image as shown in Figure, with most of the background and external illumination effects removed. The difference image is subsequently threshold. A connected component analysis is then applied to the threshold difference image to identify binary blobs that satisfy certain size and shape constraints.

### 3.1 Face (Head) Orientation Estimation:

Face (head) pose contains information about one's attention, gaze, and level of fatigue. Face pose determination is concerned with computation of the 3D face orientation and position to detect such head movements as head tilts. Frequent head tilts indicate the onset of fatigue. Furthermore, the nominal face orientation while driving is frontal. If the driver faces in the other directions (e.g., down or sideway) for an extended period of time, this is due to either fatigue or inattention.

### 3.2 Face Pose Tracking Algorithm

Given the initial face image and its pose in the first frame, the task of finding the face location and the face pose in subsequent frames can be implemented as simultaneous 3D face pose tracking and face detection. Based on the detected face pose in the previous frame, we can use the Kalman Filtering to predict the face pose in the next frame. But the prediction based [2] on Kalman Filtering assumes smooth face movements. Frequent head tilts indicate the onset of fatigue. The prediction will be off significantly if head undergoes a sudden rapid movement [2]. To handle this issue, we propose to approximate the face movement with eyes movement since eyes can be reliably detected in each frame. Then the predicted face pose is based on combining the one from Kalman Filtering with the one from eyes. The simultaneous use of Kalman Filtering and eye's motion allows performing accurate face pose prediction even under significant or rapid head movements. Details on our face pose estimation and tracking Algorithm. The proposed algorithm is tested with numerous image sequences of different people. The image sequences include a person rotating his head before an un-calibrated camera, which is approximately 1:5 meter from the person. Figure shows some tracking results under different face rotations. It is shown that the estimated pose is very visually convincing over a large range of head orientations and changing distance between the face and camera.

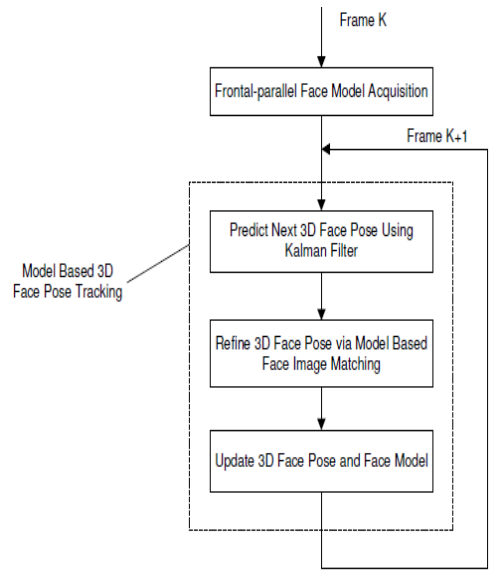

**Fig 5: Flow chart for Face Pose Tracking**

### 3.3 Eye Gaze Determination and Tracking:

Gaze has the potential to indicate a person's level of vigilance. A fatigued individual tends to have a narrow gaze. Gaze may also reveal one's needs and attention. The direction of a person's gaze is determined by two factors: the orientation of the face (face pose), and the orientation of eye (eye gaze).Face pose determines the global direction of the gaze, while eye gaze determines the local direction of the gaze. Global gaze and local gaze together determine the overall gaze of the person. So far, the most common approach for ocular-based gaze estimation is based on the determination of the relative position between pupil and the glint (cornea reaction) via a remote IR camera.

### 4. Adaptive Vehicle Speed Control Principles of Operation and Implementation:

In traditional systems, the throttle position is actuated by a mechanical link with the accelerator pedal, only operated by the driver. The amount of air flow into the engine has been adjusted by the throttle which is connected to the accelerator mechanically. Therefore, this method can't deal with the adaptive speed control and other intelligent strategy. Recently, there appeared a trend in the automotive industry of using an electronically controlled throttle valve in order to provide precise positioning of the throttle to control vehicle speed. Characteristics of these springs can be modelled as discontinuous spring in the whole operation region. Thus the block diagram of the electronic throttle control (ETC) [12] shown in Fig.6 for simplicity of the description, we use the motor angle instead of the valve angle as the control variable and consider the default angle as the origin. Thus the mechanical model is written as

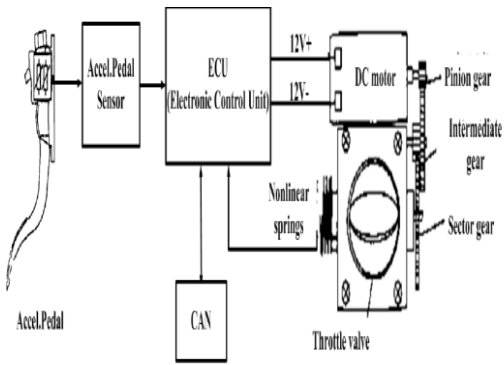

**Fig 6: General View of Electronic Throttle system**

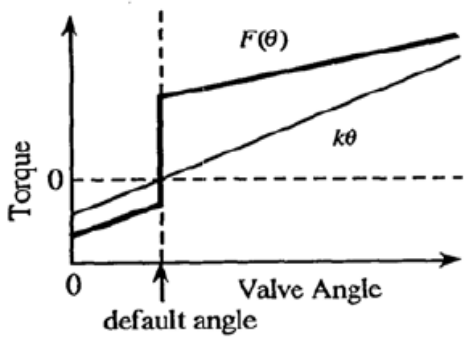

## 5. CONCLUSION

Exhaustion resulting from sleep deprivation or sleep disorders is an important factor in the creasing number of accidents. In this projected work, we have discussed the various methods of the Exhaustion techniques and we also proposed the real-time online safety prototype that controls the vehicle speed under driver fatigue. The various methods of the driver fatigue technologies are advanced system to detect fatigue symptoms in drivers and control the speed of vehicle to avoid accidents. Finally the validity of the proposed model for vehicle speed controller based on driver fatigue detection [8] is shown and the current method for fatigue detection [7] based on the Unscented Kalman Filter. Secondly, driver fatigue is confirmed and adaptive speed controller is proposed using the theory of sliding mode servo control for providing precise positioning of the throttle valve to control speed of vehicle to prevent accident.